\documentclass[10pt,twocolumn]{article}

\usepackage{3dv}
\usepackage{times}
\usepackage{epsfig}
\usepackage{graphicx}
\usepackage{amsmath}
\usepackage{amssymb}

\usepackage{comment}
\usepackage{xspace}
\usepackage{bigstrut}
\usepackage{caption}
\usepackage[caption=false,font=scriptsize]{subfig}
\usepackage{floatrow}
\usepackage{wrapfig}

\usepackage[table]{xcolor}

\usepackage[pagebackref=true,breaklinks=true,letterpaper=true,colorlinks,bookmarks=false]{hyperref}

\makeatletter
\renewcommand{\paragraph}{%
  \@startsection{paragraph}{4}%
  {\z@}{0.55ex \@plus 1ex \@minus .2ex}{-1em}%
  {\normalfont\normalsize\bfseries}%
}

\definecolor{Gray}{gray}{0.9}

\newcolumntype{a}{>{\columncolor{Gray}}c}
\newcolumntype{b}{>{\columncolor{white}}c}
\newcommand{\printfnsymbol}[1]{%
  \textsuperscript{\@fnsymbol{#1}}%
}
\newcommand{\deflen}[2]{%
    \expandafter\newlength\csname #1\endcsname
    \expandafter\setlength\csname #1\endcsname{#2}%
}

\deflen{widthdef}{0.161\textwidth}

\DeclareRobustCommand\onedot{\futurelet\@let@token\@onedot}
\def\@onedot{\ifx\@let@token.\else.\null\fi\xspace}
\def\eg{\emph{e.g}\onedot} 
\def\ie{\emph{i.e}\onedot} 
\def\cf{\emph{c.f}\onedot} 
\def\etc{\emph{etc}\onedot}

\makeatother

\threedvfinalcopy 

\def\threedvPaperID{34} 

\ifthreedvfinal\fi
\begin{document}

\title{Towards Geometry Guided Neural Relighting with Flash Photography}

\author{Di Qiu\textsuperscript{1}\thanks{Work done while at SenseTime Research}
~~~~~
Jin Zeng\textsuperscript{2}
~~~~~
Zhanghan Ke\textsuperscript{2}
~~~~~
Wenxiu Sun\textsuperscript{2}
~~~~~
Chengxi Yang\textsuperscript{2}\printfnsymbol{1}\\
\textsuperscript{1} The Chinese University of Hong Kong
~~~~~
\textsuperscript{2} SenseTime Research\\
}

\maketitle
\begin{abstract}
Previous image based relighting methods require capturing multiple images to acquire high frequency lighting effect under different lighting conditions, which needs nontrivial effort and may be unrealistic in certain practical use scenarios. 
While such approaches rely entirely on cleverly sampling the color images under different lighting conditions, little has been done to utilize geometric information that crucially influences the high-frequency features in the images, such as glossy highlight and cast shadow. 
We therefore propose a framework for image relighting from a single flash photograph with its corresponding depth map using deep learning. 
By incorporating the depth map, our approach is able to extrapolate realistic high-frequency effects under novel lighting via geometry guided image decomposition from the flashlight image, and predict the cast shadow map from the shadow-encoding transformed depth map.
Moreover, the single-image based setup greatly simplifies the data capture process. 
We experimentally validate the advantage of our geometry guided approach over state-of-the-art image-based approaches in intrinsic image decomposition and image relighting, and also demonstrate our performance on real mobile phone photo examples.
\end{abstract}
\setlength{\belowdisplayskip}{2pt} \setlength{\belowdisplayshortskip}{2pt}
\setlength{\abovedisplayskip}{2pt} \setlength{\abovedisplayshortskip}{2pt}
\setlength{\parskip}{3pt}
\setlength{\abovecaptionskip}{-5pt}
\setlength{\belowcaptionskip}{-5pt}
\setlength{\textfloatsep}{5pt}
\setlength{\intextsep}{5pt}
\setlength{\floatsep}{5pt}
\setlength{\dbltextfloatsep}{5pt}
\section{Introduction} \label{sec:intro}

Relighting has been an active research area in the past decades, with various applications including post-capture image editing, artistic visual effect, augmented virtual reality, rendering novel visualization for cultural heritages or commercial products, \etc \cite{debevec2000acquiring,sen2005dual,wang2009kernel,li2018learning,xu2018deep,zhou2019deep,kanamori2019relighting,philip2019multi}.
There are two categories of relighting methods: {\it image based relighting} and {\it reconstruction based relighting}. Previous image-based relighting approaches assume only photometric inputs and require capturing a dictionary containing hundreds or thousands of images of the scene at the same view point, where each image corresponds to a basic but distinct lighting condition such as a point or directional light source. 
Assuming the image under more complex lighting condition is a superposition of ``basis images", relighting can be done by linearly combining images in the linear intensity domain. Thus for relighting purpose, it suffices to
capture these ``basis images".
\begin{figure}[t!]
    \centering
    \includegraphics[width=1\textwidth]{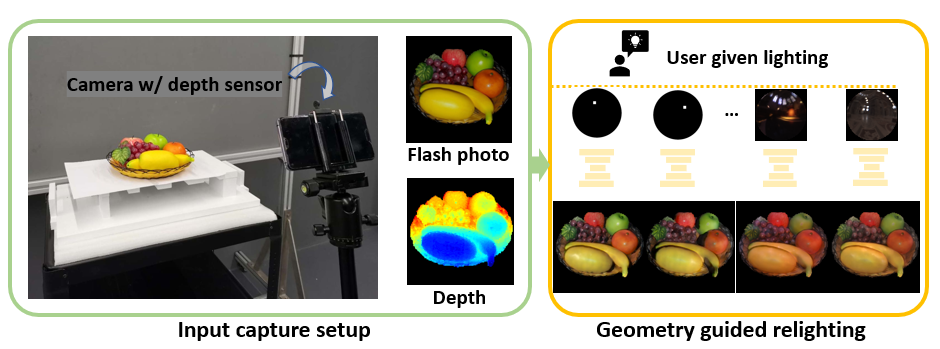}
    \caption{Illustration of our practical capture setting and the relighting application.}
    \label{fig:setting}
\end{figure}
Capturing such large amount of images with calibrated lighting (\eg using a light stage) requires considerable effort or cost, so many recent works try to ameliorate this issue by exploiting the {\it local coherence structure} of the light transport in the image domain \cite{peers2009compressive,sen2009compressive,wang2009kernel}. 

On the other hand, the reconstruction based relighting requires the complete scene information to be given, including the geometry and material properties. The relighting can be then done by Monte-Carlo ray tracing in a physically based rendering system \cite{pharr2016physically}. However, faithfully reconstructing the complete scene geometry and materials is difficult to achieve without considerable effort \cite{guo2019relightables}. 

Therefore, both relighting methodologies are impractical in the daily photo-taking scenarios, where the environment is dynamic and taking multiple measurements correspond to the same scene is not possible.
In this work we thus propose a practical approach called geometry guided image relighting. 
In terms of input, our approach {\it only} needs a flashlight photograph with its depth map, which can be easily obtained from a {\it single photo shot}. The outputs are the basic relit images under user-given directional light from the visible hemisphere, which can be further used for more complicated environmental relighting following the superposition rule. Our method {\it operates entirely in the image domain, and thus needs no explicit scene reconstruction}. Note that our input setting is easy to accommodate within many mobile phone cameras nowadays. This particular setting is illustrated in Figure \ref{fig:setting}. Our core contributions can be summarised as follows:
\vspace{-8pt}
\begin{enumerate}
    \item We propose \textit{geometry guided} intrinsic image decomposition and shadow estimation under novel directional lighting, which enables competitive performance in image decomposition and relighting tasks, validated via comprehensive experiments on two large, physically accurate synthetic datasets \cite{li2018learning,xu2018deep}.
    \vspace{2pt}
    \item Our framework requires only a flashlight image and its depth map as necessary input, which is a significantly easier setup than the previous image-based relighting methods that require calibrated lighting environment, as well as the methods that rely on multiple views for reconstruction of complex scene.
    
\end{enumerate}
\vspace{-2pt}
The rest of this paper is organized as follows. In Section \ref{sec:rel_work} we review previous work on image based relighting and surface BRDF estimation. We then explain the role of the flash photo and the depth map for the task of relighting, motivate our model design and lay out the details in Section \ref{sec:design}.
Important details about the datasets and training procedures are in Section \ref{sec:data}. Quantitative evaluation of each module on synthetic data and experiments on real data are carried out in Section \ref{sec:exp}. The paper is concluded in Section \ref{sec:conclude} with discussion on future research.

\vspace{-5pt}
\section{Related Work} \label{sec:rel_work}
\vspace{-5pt}
\paragraph{Image based relighting}
Image based relighting techniques are based on the light transport equation \cite{ng2003all}
\begin{equation} \label{eq:lte}
\mathbf{I} = \mathbf{T} \mathbf{L}
\end{equation}
where $\mathbf{I}, \mathbf{T}, \mathbf{L}$ denote the image vector, light transport matrix and the light vector, respectively, with their values all in the linear intensity domain. This equation essentially means the superposition principle for combining images under different lighting.
Earlier works \cite{debevec2000acquiring,sen2005dual} densely sample the light transport matrix in order to perform relighting. Such brute-force alike methods perform very well in general, and work with essentially arbitrary geometries and materials, but the sample size is usually huge.
Often these approaches require dedicated hardware, calibrated lighting, significant capture time and storage.  

To reduce such hard effort, recent approaches \cite{peers2009compressive,wang2009kernel,fuchs2007adaptive,sen2009compressive,ren2015image} aim to exploit the local coherence structure of the light transport by making use of the data-driven representations, reducing the required sampling to a few hundred images for a single scene. 
A very recent work \cite{xu2018deep} uses deep learning to represent such structure within large synthetic datasets, and meanwhile it learns the optimal sparse lighting directions, reducing the input to as few as five images, which is referred to as the {\it optimal sparse sampling} ({\tt OSS}) approach. However, for the above purely image-based methods, calibrated lighting condition is still essential and thus can be still unrealistic in many practical cases. 
We therefore propose to capture a single flashlight photo and its depth map to alleviate the requirement of multiple calibrated lighting.

Finally, there are also recent works that utilize the expressive power of deep neural networks for directly relighting more constrained objects like human faces from a single image \cite{sengupta2018sfsnet,sun2019single,zhou2019deep}.

\paragraph{Intrinsic image decomposition}
Estimating the {\it intrinsic images} that can be used to model an observed image is an active research area for decades. It is often called {\it inverse graphics} because they are inverse problems to the forward graphics models. The more complex the material and geometry a scene has, the more complicated the forward graphics model will become. For example, the simple shading model \cite{barrow1978recovering}
\begin{equation} \label{eq:shading}
    {\bf I} = {\bf R} \odot {\bf S}
\end{equation}
which decomposes the image ${\bf I}$ of mainly diffusive objects into its multiplicative components of reflectance ${\bf R}$ and shading ${\bf S}$. The shading term can be further decomposed into lighting configuration and surface normal. Earlier works develop optimization models which estimate the shading (and therefore surface normal) assuming the Lambertian surface property, \eg \cite{johnson2011shape} with known lighting, or \cite{barron2014shape} with unknown lighting, or with more complex materials \cite{innamorati2017decomposing}. See also references therein for more comprehensive review on this matter.
These previous works accomplish their tasks under simplifying assumptions and hand-crafted priors, thus less applicable to model general scenes. 
On the other end, the image modelling can be as complex as the entire physically based rendering system. Such is the approach of differentiable rendering \cite{david2019mitsuba}, which can potentially support a range of hard inverse graphics problems.

Material estimation itself has also been the focus of many previous works. These methods involve modelling the material, usually as a microfacet BRDF model \cite{Karis2013RealSI}. For instances,  \cite{oxholm2015shape} reconstructs the material assuming known geometry and lighting. \cite{hui2017reflectance} estimates spatially varying BRDF parameters from multiple flashlight image co-located with camera for near planar scenes. \cite{hui2016shape} performs BRDF estimation from images under different lighting using dictionary learning. Recent methods tend to utilize the representation power of deep networks trained on large datasets, \eg the very recent works \cite{deschaintre2018single,li2018materials} that reconstruct spatially varying BRDF parameters for near planar scenes. \cite{li2018learning} proposes a large synthetic dataset to learn shape, surface normal, diffuse albedo and specular roughness altogether from a single flash photograph, which is referred to as {\it single image photometric stereo}, abbreviated {\tt SIPS}.

Our relighting framework includes an intrinsic image decomposition step and follows closely the setting in \cite{li2018learning}. But our work is different in one key aspect in that we require the depth map corresponding to the flashlight image, which turns out to be very beneficial for the estimating the other intrinsic images. Note that making use of depth cue for intrinsic image decomposition tasks have been explored in \cite{barron2013intrinsic,chen2013simple} and many others. However these work consider only the simple shading model \eqref{eq:shading} and thus fall short of material reconstruction for relighting.

Finally, one can perform full reconstruction of geometry, material and environmental lighting as in \cite{wu2015appfusion} using a {\it Kinect} sensor, whose capture takes considerable time.  One can also acquire full surface reflectance as in \cite{debevec2000acquiring,Meka2019Colorgrad,guo2019relightables} using a light stage, which is in fact closely related to the image-based relighting method. While of great quality, such highly calibrated conditions is very hard to obtain and implement in common scenarios. 

%
\section{Framework}

\begin{figure*}[ht!]
    \centering
    \includegraphics[width=0.9\textwidth]{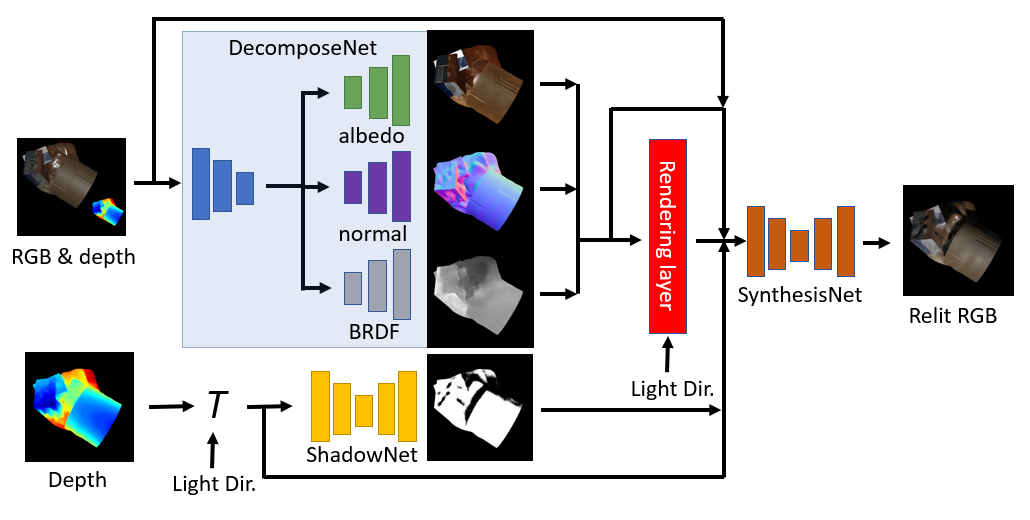}
    
    \caption{Model architecture overview. Our model can be conceptually divided into three parts: {\bf DecomposeNet}, where geometry guided intrinsic image decomposition is performed; {\bf ShadowNet}, which learns the global geometric relation between scene points after the {\it shadow encoding transform} $T$; {\bf SynthesisNet}, which re-combines the different intrinsic components and synthesizes the relit image under the user given lighting direction.}
    \label{fig:architecture}
\end{figure*}

As alluded in Section \ref{sec:intro}, the desired model must be capable of relighting with a {\it single camera shot}, which leads to our design decisions.
In terms of the framework, we have aimed for one that operates entirely in the image space, since the reconstruction can hardly be done well enough for a complex scene, which often results in unpleasant artifacts, \cf Fig.\ref{fig:recon_real}. 
In terms of the capture setup, we have required the color image to be taken under a (or almost) co-located flashlight, for three reasons:  1) the constrained lighting helps ease the complexity of scene intrinsics estimation; 2) since the flash and the camera are (almost) co-located, the flashlight image contains (almost) no shadow, thus it eliminates the hard work to distinguish shadows from textures; 3) such flashlight images are easy to obtain with many mobile devices. 
We can further assume the only light source in the input image is the flashlight, since it may be obtained by taking two photos at essentially the same time, one with flash and one without, and subtracting them in the linear intensity domain.

\subsection{Model design} \label{sec:design}


Now given a flashlight image ${I}_{\text{flash}}$ and its (possibly noisy) depth map $D$, our task is to predict the same capture but under novel, user-given lighting. To achieve this goal we incorporate geometry from the depth map that will guide our model to reason about the global high frequency component in a relit image, namely the {\it shadow} and {\it glossy highlight}. 

Since we have aimed for an image based approach, we have to compute the shadow in the image domain without resorting to reconstruction. Thus we design {\bf ShadowNet}, which learns a binary cast shadow image, where zero value indicates shadow, solely from the depth map of the flash image and the given light direction ${\omega}$. 
Note that the cast shadow is a function of the {\it global} geometry of our opaque scene. {\bf ShadowNet} will output a cast shadow map $I_{\text{shadow}}$ from the properly transformed depth map $T(\omega, D)$ such that it encodes the information of the novel lighting.
\begin{equation}
I_{\text{shadow}} = \mathbf{ShadowNet}(T(\omega, D))
\end{equation}

Besides a shadow map, we need also to learn the other non-trivial {\it local} appearance change due to the new lighting direction. This leads to our second component, {\bf DecomposeNet}, which perform principled scene analysis based on a pre-defined appearance model. Specifically, it infers the diffuse albedo map, surface normal map and spatially varying BRDF map of the captured scene. In this work we adopt the microfacet BRDF model of \cite{Karis2013RealSI}. This BRDF model is controlled by a parameter called {\it roughness}, which is the subject of our prediction. Its value ranges in $(0, 1]$, and the smaller the value, the more specular the material will appear.
\begin{equation}
I_{\text{albedo}}, I_{\text{normal}}, I_{\text{roughness}} = \mathbf{DecomposeNet}(I_{\text{flash}}, D)
\end{equation}

Since the previous components output the appearance under direct lighting, we need to add the global, indirect lighting effect to form the final image. The last component, {\bf SynthesisNet}, reconstructs the image under novel directional lighting.
\begin{equation}
\begin{aligned}
I_{\text{relit}} = \mathbf{SynthesisNet}( \omega, D, I_{\text{albedo}}, I_{\text{normal}},  \\
                  I_{\text{roughness}}, I_{\text{shadow}},
                  I_{\text{flash}}) 
\end{aligned}
\end{equation}
The structure of our model is illustrated in Figure \ref{fig:architecture}. We next explain in details the design principle and loss function of each individual component. More details can be found in the supplementary material.

\paragraph{Learning cast shadow}
One of the most significant factors that contributes to the visual realism of a synthesized image is cast shadow. 
There are many ways to encode the visibility information and we choose the following method that works well in practice. 
We first convert the depth image $D$ to the corresponding point cloud image, namely a three-channel image with each channel stores the $x, y, z$ coordinates in the camera's Euclidean coordinate system.
Suppose the light is in direction $\omega$. We complete it into an orthonormal basis, in the form of a matrix $R$ such that the third column is $\omega$. We then transform a point $p=(x,y,z)^T$ by
\begin{equation}
T:p \mapsto R^T p + t
\end{equation}
where $t$ is a suitable translation. In our case we simply use $t=(0,0,1)^T$. Note that this transform can be implemented as a convolution layer of kernel size $1\times1$. 


The transformed point coordinates, which is again a three-channel image, will be the input of an hour-glass shaped convolutional neural network with skip links (U-Net) to predict a shadow image.
We use the pixel-wise binary cross-entropy loss $\mathcal{L}_{\text{shadow}}$ for supervised training. More details about its design motivation and computation of $R$ can be found in the supplementary material.

\paragraph{Learning image decomposition}
Having described how to generate the shadow image, which is about learning the geometric relations between scene points that are possibly far apart, now we head to the more local image analysis task of intrinsic image decomposition, making use of the flashlight color image. 

Since the three tasks, namely estimating diffuse albedo, surface normal and roughness, are not mutually exclusive, we use a shared encoder that branches into three different decoders, as shown in part of Figure \ref{fig:architecture}. 



The shared encoder then takes $I_{\text{flash}}, P$ as inputs, and the feature from the last layer are fed into three individual decoders, with skip-links connected to intermediate layers of the encoder to ensure that the high-frequency details are preserved. It is worth mentioning that, although we assumed the depth image in the input, surface normal estimation will benefit from the flashlight color image. One obvious reason is that the depth map obtained may contain noise. But the benefit even holds true for synthetic data with perfect depth, since the depth map only represents the large-scale smooth geometry, while the \textit{micro-scale non-smooth} geometry are represented separately using surface normal textures \cite{blinn1978simulation}. 

The roughness prediction branch is supervised by pixel-wise binary cross-entropy loss $\mathcal{L}_{\text{roughness}}$. The normal prediction branch and the diffuse albedo prediction branch use $L^1$ loss on both pixel value and image gradients, denoted by $\mathcal{L}_{\text{normal}}$ and $\mathcal{L}_{\text{albedo}}$.

\paragraph{Image synthesis under novel lighting}

Once we have the diffuse albedo, surface normal and roughness maps, we can already compute in close form a $1$-bounce image $I_{\text{render}}$ of the imaged scene under light from arbitrary direction $\omega$, though {\it without cast shadow}. 
To add the cast shadow, enhance the quality, we pass the concatenation of $I_{\text{shadow}}$, $I_{\text{render}}$, $I_{\text{flash}}$, $I_{\text{albedo}}$, $I_{\text{normal}}$, $I_{\text{roughness}}$ and  $T(\omega, D)$ to a final U-Net structure {\bf SynthesisNet} to predict the final image under lighting from direction $\omega$. This network is also trained using a combination of $L^{1}$ losses on both image and image gradient, denoted by $\mathcal{L}_{\text{relight}}$.

\subsection{Datasets and training procedures} \label{sec:data}
It is often essential to have a large amount of high quality data sampled from diverse scenarios for a deep learning method to work well.
For the image relighting task, there are usually hundreds or thousands of ``basis images" that come with a single view of a scene.
Consequently, to collect a large dataset containing diverse scenes is very challenging. 
To address this issue, Xu {\it et al.} \cite{xu2018deep} proposed a large photo-realistic synthetic dataset. The major part of Xu's dataset contains 500 training scenes and 100 testing scenes, each generated by taking random combinations of shapes, surface normal maps, diffuse albedo textures and specular roughness maps. For each scene, a total of 1053 images are rendered with directional lighting over the visible hemisphere. We consider Xu's dataset to be one of the ideal datasets for training the relighting model with directional lighting. To adapt Xu's dataset to our task, we additionally render the depth map for each scene, using the geometry provided in their dataset, as well as the binary shadow image for each lighting direction. 

Unfortunately, Xu's dataset does NOT contain those ground truth surface normal maps, diffuse albedo textures and specular roughness maps.
This means that we cannot directly supervise our {\bf DecomposeNet} using Xu's dataset alone. We address this problem by utilize another complimentary dataset proposed by Li {\it et al.} \cite{li2018learning}. It contains ground truth intrinsic images but not the relighting, and a different BRDF model than \cite{xu2018deep}.
To train our model with these two datasets, we adopt a stage-wise training strategy as follows. 

In the first stage, we train the module {\bf ShadowNet} on Xu's data using loss $\mathcal{L}_{\text{shadow}}$, and {\bf DecomposeNet} on Li's data using loss
$\mathcal{L}_{\text{albedo}} + 
\mathcal{L}_{\text{normal}} + 
\mathcal{L}_{\text{roughness}} $,
both for 5 epochs. The training on two datasets can be done in parallel. 
In the second stage, we fix the pretrained networks {\bf ShadowNet} and {\bf DecomposeNet} from the first stage and train {\bf SynthesisNet} on Xu's data using $\mathcal{L}_{\text{relight}}$ for 5 epochs. For all training, the batch size is set to be $4$, and we use the Adam optimizer with default hyper-parameters with initial learning rate $5\times 10^{-4}$ and decay it by $\times0.1$ every two epochs. Totalling at $27M$ parameters, our model is relatively more compact in sizes than the models released by \cite{xu2018deep} ($33M$) and \cite{li2018learning} ($64M$). We will open source our implementation to promote future research in this direction.

\section{Evaluation and results}

\label{sec:exp}

\subsection{Influence by the quality of depth input}
\begin{figure*}[!t]
    \scriptsize
    
    \subfloat{\includegraphics[width=0.98\widthdef]{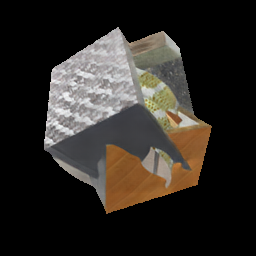}} 
   \hspace{0.1pt}
    \subfloat{\includegraphics[width=0.98\widthdef]{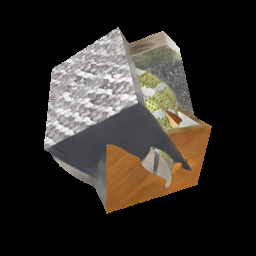}}
   \hspace{0.1pt}
    \subfloat{\includegraphics[width=0.98\widthdef]{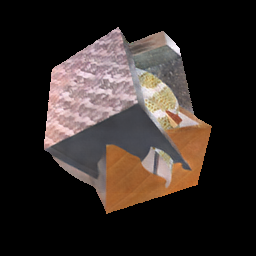}}
    \hspace{0.1pt}
    \subfloat{\includegraphics[width=0.98\widthdef]{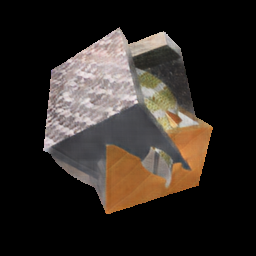}}
   \hspace{0.1pt}
    \subfloat{\includegraphics[width=0.98\widthdef]{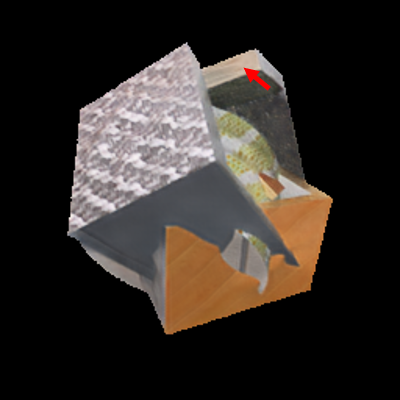}}
   \hspace{0.1pt}
   \subfloat{\includegraphics[width=0.98\widthdef]{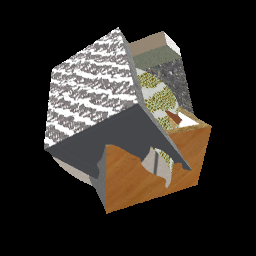}}\\[-2.5ex]
    \subfloat{\includegraphics[width=0.98\widthdef]{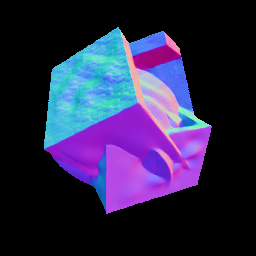}} 
   \hspace{0.1pt}
    \subfloat{\includegraphics[width=0.98\widthdef]{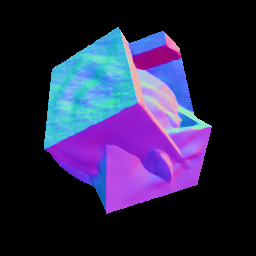}}
   \hspace{0.1pt}
    \subfloat{\includegraphics[width=0.98\widthdef]{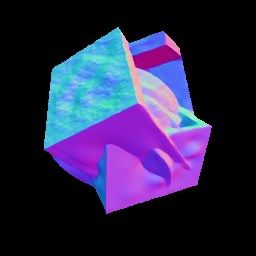}}
    \hspace{0.1pt}
    \subfloat{\includegraphics[width=0.98\widthdef]{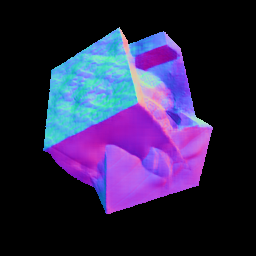}}
   \hspace{0.1pt}
    \subfloat{\includegraphics[width=0.98\widthdef]{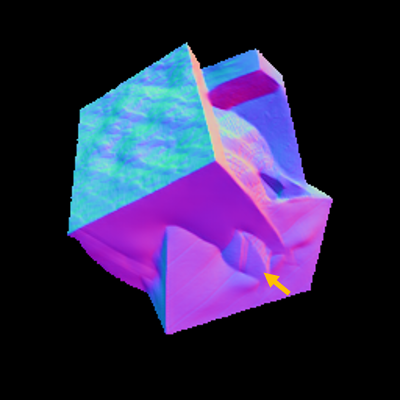}}
   \hspace{0.4pt}
    \subfloat{\includegraphics[width=0.98\widthdef]{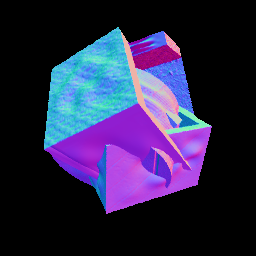}}\\[-2.5ex]
    \addtocounter{subfigure}{-12} 
    \subfloat[{\tt Clean.} ]{\includegraphics[width=0.98\widthdef]{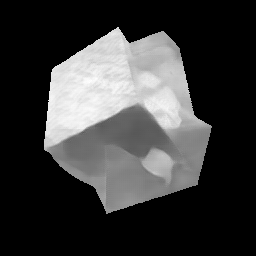}}
    \hspace{0.1pt} 
    \subfloat[{\tt Noisy}]{\includegraphics[width=0.98\widthdef]{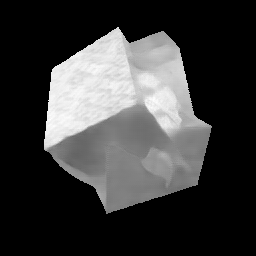}}
    \hspace{0.1pt}
    \subfloat[{\tt Env.}]{\includegraphics[width=0.98\widthdef]{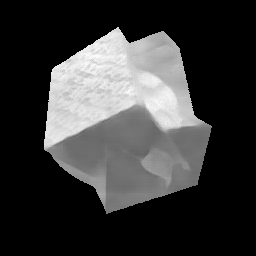}}
    \hspace{0.1pt} 
    \subfloat[\cite{li2018learning}{\tt Init.}]{\includegraphics[width=0.98\widthdef]{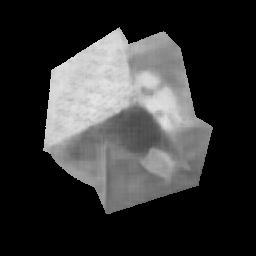}}
    \hspace{0.1pt}
    \subfloat[
    \cite{li2018learning}{\tt Refine.}]{\includegraphics[width=0.98\widthdef]{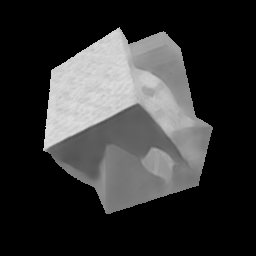}}
    \hspace{0.1pt}
    \subfloat[Truth]{\includegraphics[width=0.98\widthdef]{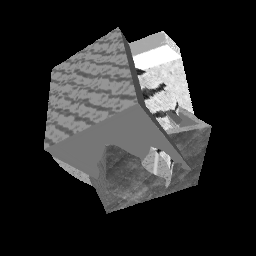}}
    
    \caption{Visual comparison for {\bf DecomposeNet} under different input configurations on Li's testing dataset \cite{li2018learning}. Each row shows albedo, normal and roughness map respectively. The initial and final results from the three-stage model of \cite{li2018learning} are also included for comparison. Note our albedo and normal estimation are much more accurate than \cite{li2018learning} thanks to the geometry guidance. Note the arrows marked in (c). Please zoom-in for better details.}
    \label{fig:li_compare}
\end{figure*}

\begin{figure*}[t!]
    \centering
    \includegraphics[width=0.97\textwidth]{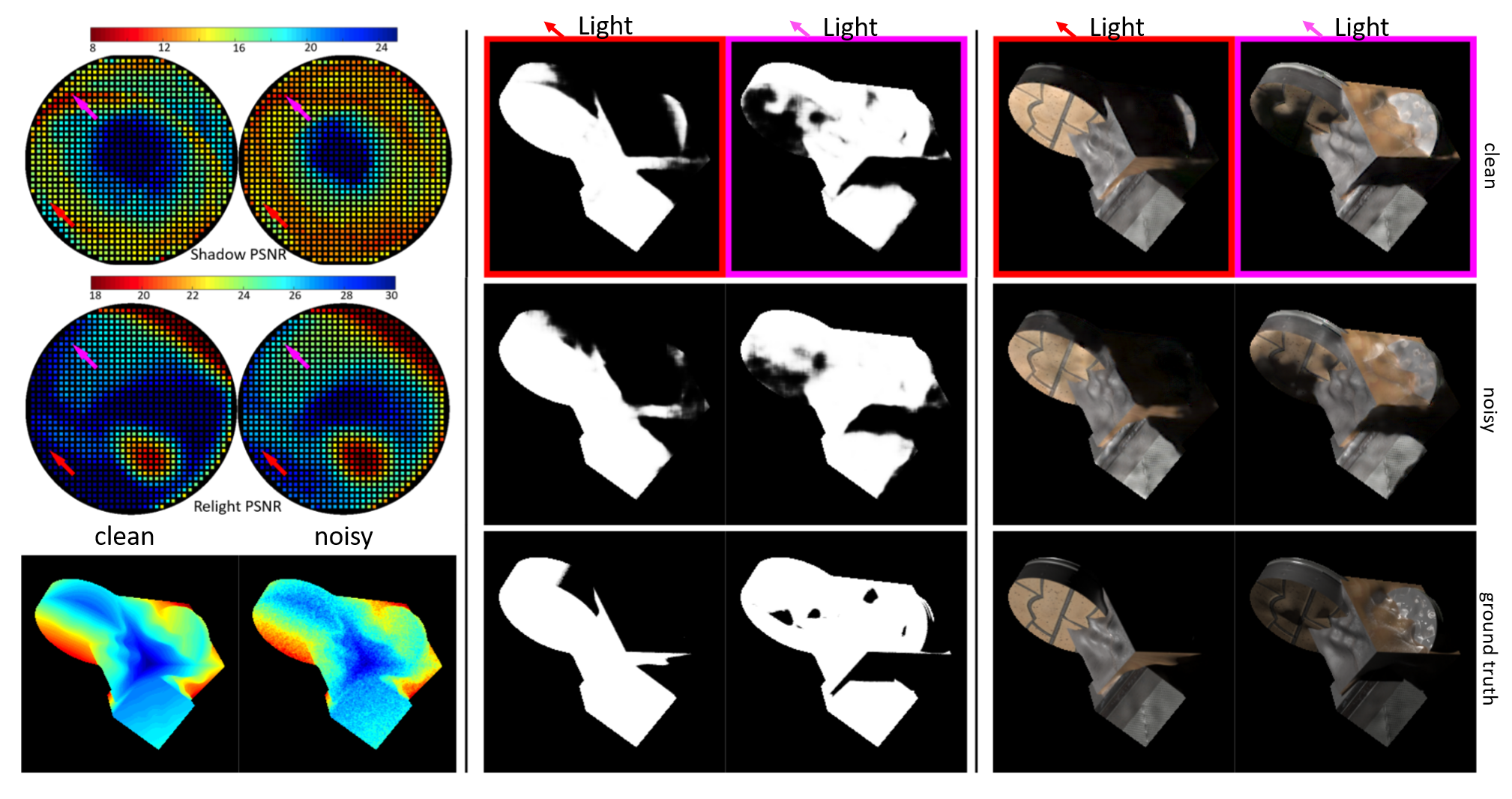}

    \caption{Visual comparison for {\bf ShadowNet} and {\bf SynthesisNet} trained with clean and noisy depths. On the left we show the PSNR-direction plot for the current scene, with clean and noisy depth inputs illustrated at the bottom. The shadow estimation and relighting under novel light directions (marked on the PSNR plot) are shown in the middle and right columns. The first row represents results of {\tt Clean}, and second row {\tt Noisy}, and the last row ground-truth.}
    \label{fig:clean_noisy}
\end{figure*}

\begin{figure}[h!]
    \centering
    \subfloat{\includegraphics[width=0.8\textwidth]{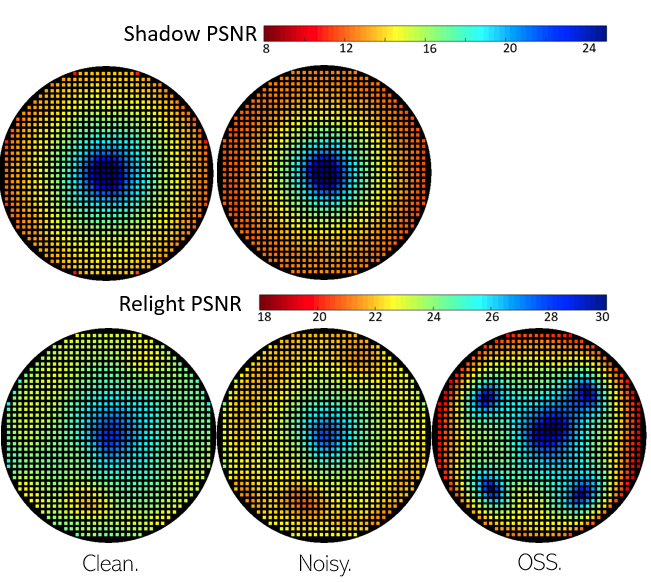}}
    \caption{Average PSNRs with respect to lighting directions for both {\bf ShadowNet} and {\bf SynthesisNet} on Xu's testing dataset \cite{xu2018deep}. The result from {\tt OSS} in the default setting (five input images under different calibrated directional lighting) is included for comparison.}
    \label{fig:loss}
\end{figure}

Since our framework requires depth map as input, a first question would be how the quality of depth affects the results, in terms of {\it both} training and testing, since for real data it is difficult to obtain perfect depth as ground-truth. 
In this respect, we quantitatively evaluate our {\bf ShadowNet}, {\bf DecomposeNet} and {\bf SynthesisNet} modules. We train two models using identical procedures, one ({\tt Clean}) with clean depths and the other ({\tt Noisy}) with random Gaussian noise and Gaussian blur applied. For input depth in range $[0,1]$ of size $256\times256$, we choose $\mu=0,\sigma=6.25\times10^{-2}$ for Gaussian noise and $\sigma=1$ for Gaussian blur. An instance for clean and noisy depths is illustrated in the bottom-left of Fig.\ref{fig:clean_noisy}.
During testing, {\tt Clean} uses original testing data and {\tt Noisy} uses the same except for the noisy depth under the same perturbation scheme.  
The numerical results in Mean Squared Error (MSE) are shown in Table \ref{tb:clean_noisy} with PSNR visualization for each testing lighting direction in Fig.\ref{fig:loss}. Typical testing results are shown in Fig.\ref{fig:li_compare} and \ref{fig:clean_noisy}.

From Table \ref{tb:clean_noisy}, we can observe that for {\bf DecomposeNet}, the albedo and BRDF roughness estimation tend to be more robust to the noise in depth. In fact, the albedo task even sees an increase in performance. The increase in performance of diffuse albedo estimation is possibly due the nature of the task. The diffuse appearance of the flash image is less affected by local depth fluctuations and thus noisy geometry can be treated as an effective data augmentation. Note this is not the case for the normal estimation and roughness estimation.
The {\bf ShadowNet} also has a drop in performance and thus affects {\bf SynthesisNet}. This is expected, since shadow map prediction should rely on accurate geometry. Note that all results from {\bf DecomposeNet} compare favorably to {\tt SIPS} \cite{li2018learning}, which is a considerably more complicated model. This demonstrates the effectiveness of geometry guided image decomposition.

\vspace{15pt}
\begin{table} [h!]
  \centering \footnotesize
   \setlength{\tabcolsep}{0.5pt}
   \resizebox{\columnwidth}{!}{
    \begin{tabular}{c|ccccc}
    \hline
    MSE  & Albedo & Normal & Roughness & Shadow & Relight  \bigstrut\\
    \hline
    \rowcolor{Gray} {\tt Clean.} &   $0.72\times10^{-2}$ &  $\mathbf{1.53\times10^{-2}}$   & $\mathbf{4.18\times10^{-2}}$     &   $\mathbf{1.26\times 10^{-2}}$ &
    $\mathbf{0.40\times 10^{-2}}$
    \bigstrut[t]\\
    \hline
    {\tt Noisy.}&     $\mathbf{0.48\times10^{-2}}$    &   $1.85\times10^{-2}$   &  
    $4.23\times 10^{-2}$     &    
    $5.53\times 10^{-2}$  &
    $0.54\times 10^{-2}$
    \bigstrut[t]\\
    \hline
        {\cite{li2018learning}}&     ${4.84\times10^{-2}}$  &  ${3.81\times10^{-2}}$   & ${1.93\times10^{-1}}$    &    
    $N/A$  &
    $N/A$
    \bigstrut[t]\\
    \hline
    \end{tabular}%
    }
    \caption{Influence by the quality of depth input. The quality of depth has a notable effect on normal and shadow estimation.}
    \label{tb:clean_noisy}
\end{table}%
\begin{figure}[h!]
    \centering
    \subfloat{\includegraphics[width=0.95\widthdef]{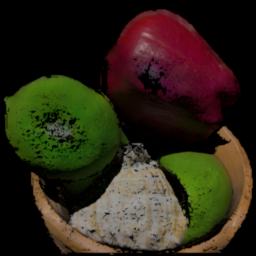}}
    \hspace{1pt}
    \subfloat{\includegraphics[width=0.95\widthdef]{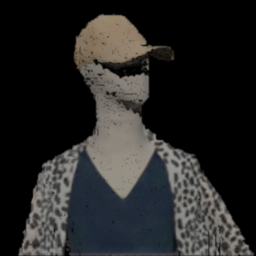}}
    \hspace{1pt}
    \subfloat{\includegraphics[width=0.95\widthdef]{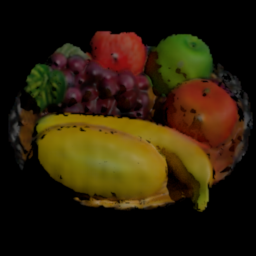}}
    \caption{Reconstruction based relighting results. Note the sever artifacts due to the difficulty of generating a good mesh from a single noisy depth for complex shapes.  For comparison, our results are shown in Fig.\ref{fig:real}}
    \label{fig:recon_real}
\end{figure}

\vspace{-5pt}

\begin{figure*}[t!]
    \centering
    \subfloat{\includegraphics[width=1.01\widthdef]{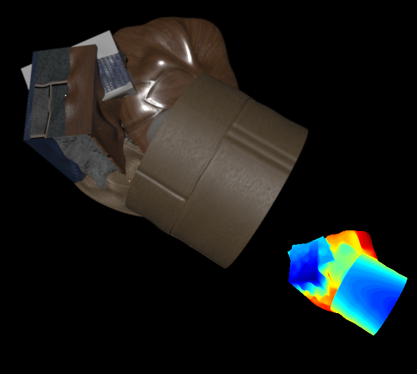}}
    \hspace{0.3pt} 
    \subfloat{\includegraphics[width=\widthdef]{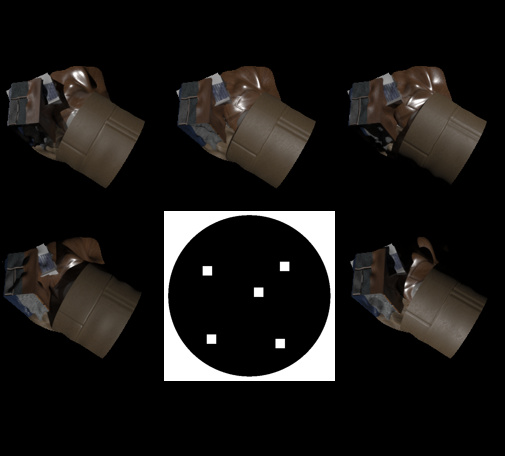}}
    \hspace{1pt} 
    \subfloat{\includegraphics[width=\widthdef]{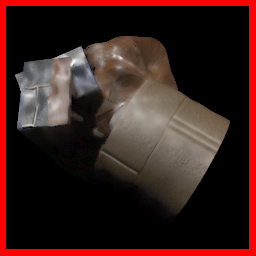}}
    \subfloat{\includegraphics[width=\widthdef]{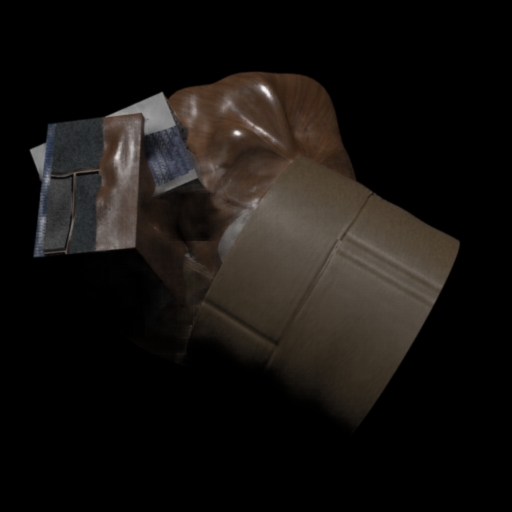}}
    \subfloat{\includegraphics[width=\widthdef]{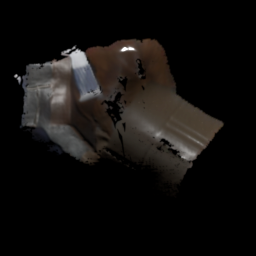}}
    \hspace{0.5pt}
    \subfloat{\includegraphics[width=\widthdef]{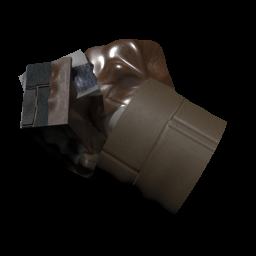}}
    \\[-1.8ex]
    \addtocounter{subfigure}{-6}
    \hspace{0.2\widthdef}
    \subfloat[Ours  vs. {\tt OSS}\cite{xu2018deep} ]{\includegraphics[width=1.6\widthdef]{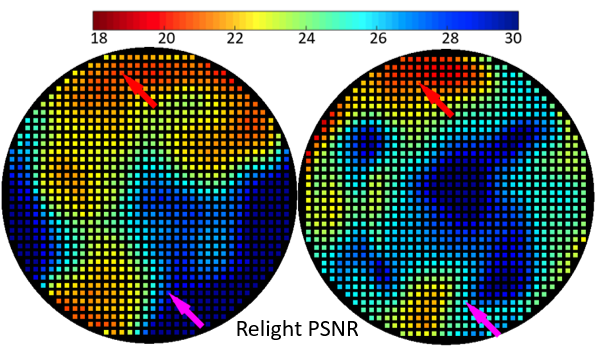}}
    \hspace{0.21\widthdef}
    \subfloat[Ours ]{\includegraphics[width=\widthdef]{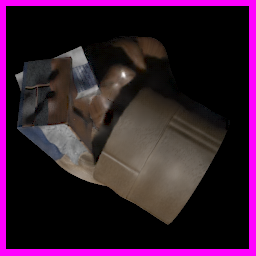}}
    \subfloat[{\tt OSS} \cite{xu2018deep}]{\includegraphics[width=\widthdef]{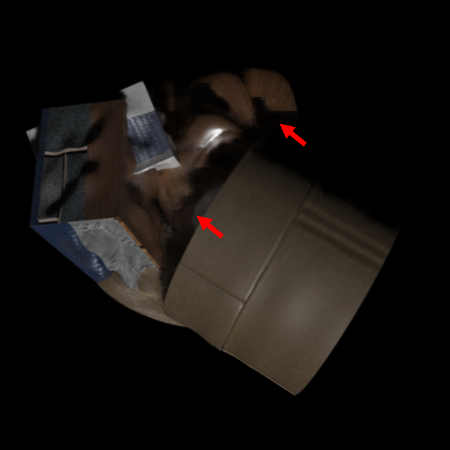}}
    \subfloat[{\tt SIPS} \cite{li2018learning} ]{\includegraphics[width=\widthdef]{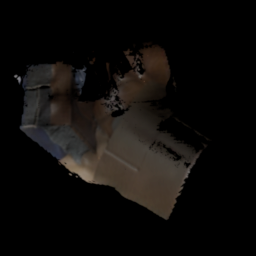}}
    \hspace{0.5pt}
    \subfloat[Truth]{\includegraphics[width=\widthdef]{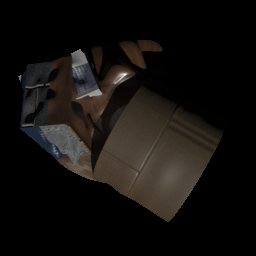}}
    
    \caption{Visual comparison of relighting performance. The input setting between ours and {\tt OSS} is illustrated on the top-left, and the PSNR-direction plot on the bottom-left. Two {\it large angle} lighting directions are chosen for visualization. Please zoom-in for better details.}
    \label{fig:xucompare1}
\end{figure*}

\deflen{envfig}{70pt}
\begin{figure}[h!]
    \centering
    \subfloat{\includegraphics[width=\envfig]{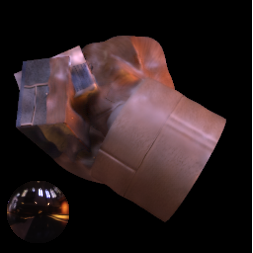}}
    \hspace{1pt} 
    \subfloat{\includegraphics[width=\envfig]{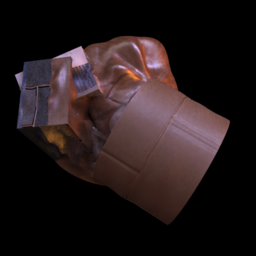}}
    \hspace{1pt} 
    \subfloat{\includegraphics[width=\envfig]{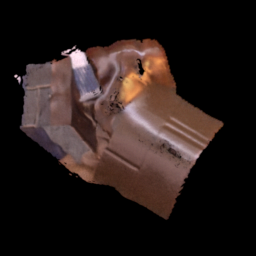}}
    \\[-2ex]
    \addtocounter{subfigure}{-3}
    \subfloat[Ours]{\includegraphics[width=\envfig]{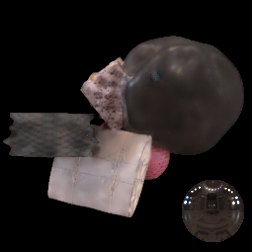}}
    \hspace{1pt} 
    \subfloat[{\tt OSS} \cite{xu2018deep}]{\includegraphics[width=\envfig]{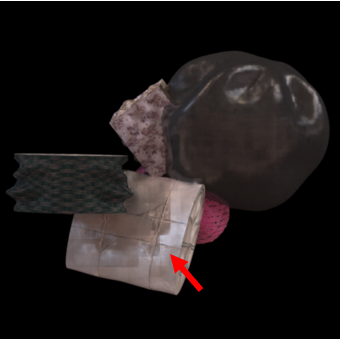}}
    \hspace{1pt} 
    \subfloat[{\tt SIPS} \cite{li2018learning} ]{\includegraphics[width=\envfig]{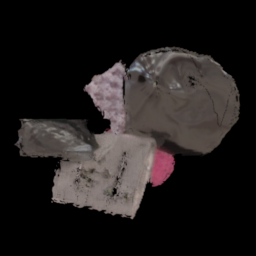}}
    \caption{Comparison for relighting under novel environmental light. Note the block-ish artifact from {\tt OSS} in the second row (b), because it makes no use of geometric information. {\tt SIPS} reconstruction does not work well when the underlying scene geometry is complex, along with other technical problems during the scene reconstruction. Please zoom-in for better details.}
    \label{fig:xucompare2}
\end{figure}

\begin{figure*}[t!]
    \centering
    \subfloat{\includegraphics[width=0.95\widthdef]{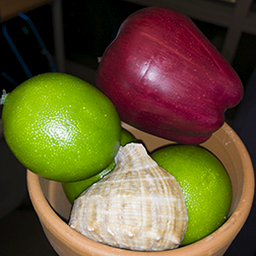}}
    \hspace{1pt} 
    \subfloat{\includegraphics[width=0.95\widthdef]{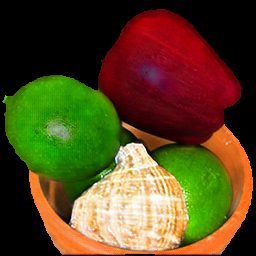}}
    \hspace{1pt} 
    \subfloat{\includegraphics[width=0.95\widthdef]{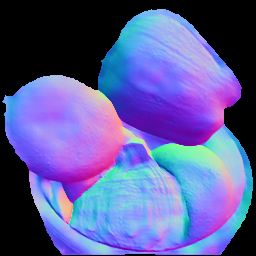}}
    \hspace{1pt}
    \subfloat{\includegraphics[width=0.95\widthdef]{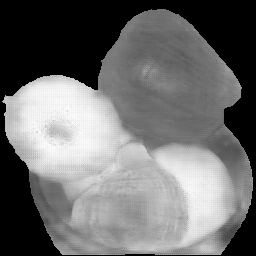}}
    \hspace{1pt} 
    \subfloat{\includegraphics[width=0.95\widthdef]{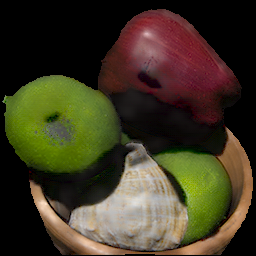}}
    \hspace{1pt} 
    \subfloat{\includegraphics[width=0.95\widthdef]{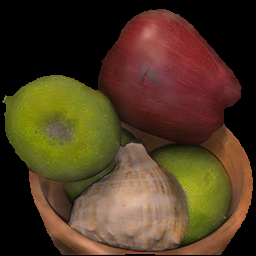}}
    \\[-2ex]
    \subfloat{\includegraphics[width=0.95\widthdef]{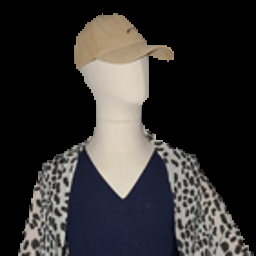}}
    \hspace{1pt} 
    \subfloat{\includegraphics[width=0.95\widthdef]{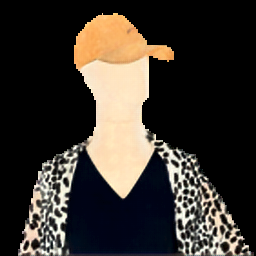}}
    \hspace{1pt} 
    \subfloat{\includegraphics[width=0.95\widthdef]{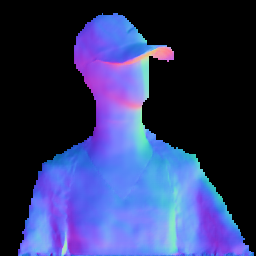}}
    \hspace{1pt}
    \subfloat{\includegraphics[width=0.95\widthdef]{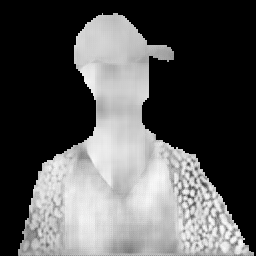}}
    \hspace{1pt} 
    \subfloat{\includegraphics[width=0.95\widthdef]{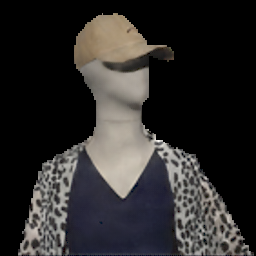}}
    \hspace{1pt} 
    \subfloat{\includegraphics[width=0.95\widthdef]{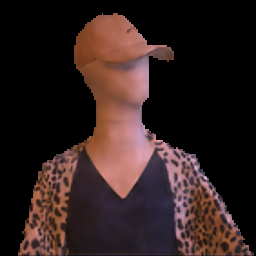}}
    \\[-2ex]
    \addtocounter{subfigure}{-12}
    \subfloat[Flash]{\includegraphics[width=0.95\widthdef]{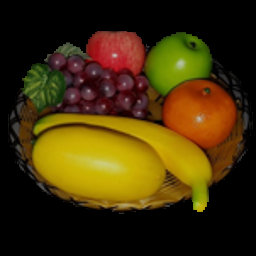}}
    \hspace{1pt} 
    \subfloat[Albedo]{\includegraphics[width=0.95\widthdef]{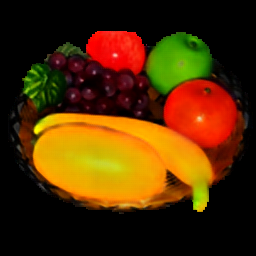}}
    \hspace{1pt} 
    \subfloat[Normal]{\includegraphics[width=0.95\widthdef]{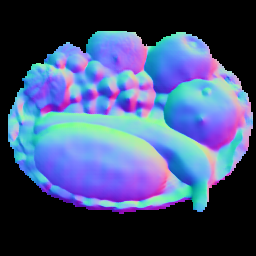}}
    \hspace{1pt}
    \subfloat[Roughness]{\includegraphics[width=0.95\widthdef]{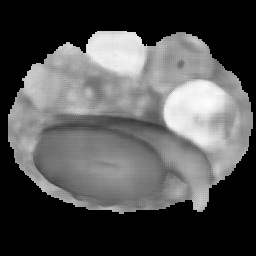}}
    \hspace{1pt}
    \subfloat[Dir. light]{\includegraphics[width=0.95\widthdef]{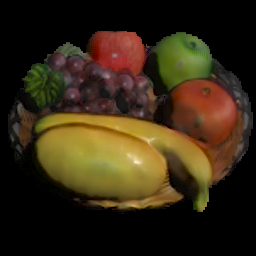}}
    \hspace{1pt} 
    \subfloat[Env. light]{\includegraphics[width=0.95\widthdef]{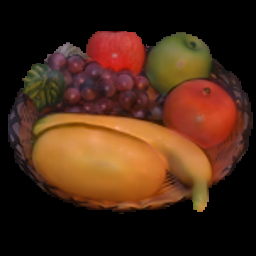}}
    
    \caption{Visual evaluation of our model on real data under novel directional light and environmental light. The image and depth in the first row is taken from the real dataset provided in \cite{li2018learning}. The images and depths in the last two rows are taken by our mobile phone camera with Time-of-Flight sensor.}
    \label{fig:real}
\end{figure*}

\subsection{Comparison with reconstruction based method} \label{sec: compare_recon}
For this baseline approach we reconstruct the scene using {\it MeshLab} \cite{cignoni2008meshlab} from the input noisy depth map. Specifically, we choose the ball-pivoting method \cite{bernardini1999ball} which yields the best empirical meshing result. We texture map our inferred albedo, normal and material images, and relight the scene via physically based rendering. The results on our real data captured using a mobile phone are shown in Fig.\ref{fig:recon_real}. As can be seen, there are severe artifacts in the cast shadows and holes in the surface, since it is difficult to produce a nice mesh from a single noisy depth map of a complex scene. In contrast, we show our results under the same lighting in Fig.\ref{fig:real} where such artifacts are effectively avoided.  This should justify our use of image-based approach.

\subsection{Relighting performance evaluation}
To conduct quantitative evaluation, ground truth relighting results are necessary. Since most real datasets do not contain geometry that fits our input setting, we therefore utilize photo-realistic synthetic data for this purpose, following the practice of \cite{xu2018deep}. The main state-of-the-art relighting approach to compare is {\it optimal sparse sampling} ({\tt OSS}) \cite{xu2018deep}, which in its default setting utilizes {\it five} input images, each under {\it calibrated} directional light source. Note that its capture setting is much more complicated than ours.
We compare the input setting of {\tt OSS} with ours in the top-right corner of Fig.\ref{fig:xucompare1}. We visualize results of the three approaches under the {\it large angle} lighting in Fig.\ref{fig:xucompare1}.  The average PSNR over the Xu's testing dataset of each direction is shown in Fig.\ref{fig:loss}. We also compare the environmental relighting performance qualitatively with {\tt OSS} and visualize it in Fig.\ref{fig:xucompare2}. More comparison can be found in the supplementary video.

Quantitatively, as shown in Fig.\ref{fig:loss}, our relighting results compare favourably with {\tt OSS} for small lighting angels and is satisfactory when the angle is large. We observe that {\tt OSS} has more block-ish artifacts in shadow, as marked by red arrows in the second row of Fig.\ref{fig:xucompare1}(c), and the second row of Fig.\ref{fig:xucompare2}(b), since geometric information is absent from the shadow synthesis in {\tt OSS}.
Thanks to multiple samples, {\tt OSS} is better at producing more faithful highlights. Note that our highlight computation uses the original BRDF as in Li's dataset \cite{li2018learning}.
Since the {\tt OSS} has different BRDFs with no ground truth given, our rendering layer computes the rendering that is incompatible with {\tt OSS}. {\bf SynthesisNet} can compensate for this defect to a certain extent, though not able to perfectly close the gap, causing blurry highlight in the final results under large angle lighting.
Better performance can be expected if a unified dataset is available. 



We also consider the related approach {\tt SIPS} based on {\it scene reconstruction} from a single flashlight image \cite{li2018learning}, where the depth estimated from the color image and not taken into account for guiding the decomposition task. The depth estimation of this method usually degrades severely when the scene is complex and the relighting completely fails. Here we show the qualitative results in Fig.\ref{fig:xucompare1} and Fig.\ref{fig:xucompare2} using the same reconstruction pipeline detailed in Section \ref{sec: compare_recon}.
We can observe that {\tt SIPS} behaves poorly in the relighting task for the complex shapes in Xu's dataset, and the associated technical problems \eg holes, from the scene reconstruction further degrades the visual quality. This demonstrates the advantage of our image based method without need for reconstruction.

Finally, in Fig.\ref{fig:real} we demonstrate some results by directly applying our trained model on real images captured by mobile phone cameras. Notice the realistic cast shadows we produce between the individual objects as well as the glossy highlights under novel directional and environmental lighting. Please refer to our supplementary video for more thorough visualization results.

\section{Conclusion and future work} \label{sec:conclude}

We have proposed a framework for geometry guided neural image relighting with flash photography. The flashlight provides us a simple constrained lighting condition for material inference, and the geometry guidance from the depth map allows our model to reason about the global image appearance under a novel lighting condition. Our modular design allows the model to learn different high frequency effects like cast shadows and glossy highlights separately, while remains conceptually simple. 

There are several future directions which can be explored. The first would be extending this framework to the multi-modal sensor domain, for example a Time-of-Flight(ToF) RGB-D camera module. In this case 
the infrared image may provide additional cues for material inference, and can potentially further alleviate the flashlight constraint.
Secondly, creating a physically accurate synthetic dataset with full lighting conditions and ground-truth material and geometries, as well as large real dataset with full lighting conditions and geometry, can be extremely helpful to promote future research in this direction.


{\small
\bibliographystyle{ieee}
\bibliography{reference}
}

\appendix
\section{Additional explanation on our model design}
\subsection{Why we need shadow estimation and intrinsic image decomposition} \label{sec:prelim}
To motivate our model design, we need first introduce some elements of the physically based rendering model for an opaque scene.
Let $L(p'\to p)$ denote the radiance from point $p'$ to $p$ in an opaque scene $\mathcal{A}$. Its resulting exiting radiance $L(p \to p'')$ from $p$ to another $p''$ can be computed as a certain fraction of $L(p'\to p)$ \cite{pharr2016physically}:
\begin{equation} \label{eq:brdf} 
L(p \to p'') = f(p'\to p \to p'')G(p'\to p)L(p'\to p)
\end{equation}
There are two distinctive factors in the above expression. 
The first is $f(p' \to p \to p'')$, the BRDF at $p$, which encodes the material property.
Such a BRDF can be decomposed into diffusive and glossy components. The diffusive component (which is equivalent to the diffuse albedo) is simply the constant term in a BRDF. Hence the more glossy a BRDF is, the more it depends the geometric relation between the points $p', p$ and $p''$. The second $G(p' \to p)$ term encodes the geometric relation between $p'$ and $p$, including global intensity fall-off related to the distance between $p'$ and $p$; Lambert's cosine law, which concerns the surface normals at $p$ and $p'$; most importantly, their {\it mutual visibility}, which is a dominating term in $G$ when the scale of the scene is small. 

Now in Expression \eqref{eq:brdf}, consider the source light ray $p' \to p$ to be in direction $\omega$,  and $p''$ is our camera. This equation corresponds to the $1$-bounce (\ie direct illumination) image under directional lighting. The outgoing radiance $L(p \to p'')$ will vary smoothly with respect to $\omega$, {\it except} when the mutual visibility term changes its value, or enters/exits the glossy component of the BRDF.
In the image space, the first kind of discontinuity appears in the form of {\it cast shadow}, and the latter appears in the form of {\it glossy highlights}. These discontinuities are the main source of {\it local incoherence} in light transport. Hence for relighting, we argue that predicting them will be {\bf necessary} to alleviate multiple captures under calibrated lighting. A naive model without factoring the relighting into these components exhibit very poor results on complex scenes.


If the geometry of the scene is available, it is possible to have a more reliable estimate of the BRDF $f$ and the geometric term $G$. 
First of all, material inference can be made under particular profile of intensity, light direction and surface local geometry, which makes the challenging problem of material estimation easier to solve than purely based on color images \cite{li2018learning}. 
Also, the mutual visibility term can also be estimated by reasoning the geometric relations {\it between} the scene points. Thus our model first performs analysis on both image and geometry to obtain estimation on the BRDF and visibility terms, which result in the glossy highlight and cast shadow in an image under novel lighting.

\subsection{Why encoding cast shadow as view-dependent scene coordinates}
In computer graphics, the computation of the cast shadow map has been successful with well-modeled geometry. The main computation is to check whether the ray between a scene point and the light source is blocked by some other opaque scene surface. In screen space rendering, checking whether the {\it pixel} $(i,j)$ is in shadow is done by comparing the smallest depth value rendered from the {\it light perspective}, called {\it shadow mapping} (which requires a new rendering pass), with the true depth value at $(i,j)$. If the shadow mapping's value is smaller, then the pixel should be in shadow.


However, shadow computation has been less explored with a single depth map without resorting to scene reconstruction, which is often a challenging task in application scenarios. 
Here we propose to directly learn the shadow image from depth map and the given light direction $\omega$. The key to our approach is a transform $T$ of the depth map defined by $\omega$. The principle is the same with that of the screen space method: the shadow created by a directional light is the same with the occlusion seen from an orthographic camera from the light direction. So we can encode this shadow information by transforming the scene points from the camera's Euclidean coordinate system to the light's Euclidean coordinate system. We then use a neural network to retrieve the encoded shadow image.
The computation of matrix $R$ can be done as follows. Given the vector $\omega$, we find two basis vectors $e_i, e_j$ out of the standard coordinate frame to from $(e_i, e_j, \omega)$ so that they are linearly independent. Then we apply the {\it Gramm-Schmidt orthogonalization} procedure to form the matrix $R$. In our implementation, we simply use the python provided function {\it linalg.null\_space}. Although it may be the case for nearby directions different $R$'s are generated, we find it has little impact to the results and can be treated as a form of data augmentation.

\section{Network details}
We use the Leaky-ReLU non-linearity with negative slope $-0.1$. All convolutional parameters are initialized using He's method \cite{he2015delving}, the bias terms are initialized to be zero. Albedo and relighting prediction layers has no non-linearity and no bias term; normal prediction layer has per pixel normalization to have unit Euclidean norm; roughness and shadow prediction layers have sigmoid activation. The specific structure are detailed in the following table.
\begin{table}[htbp]
\footnotesize
  \centering
  \renewcommand{\arraystretch}{1}
   \begin{tabular}{|m{29pt}<{\centering}|m{8pt}<{\centering}m{3pt}<{\centering}m{27pt}<{\centering}|m{5pt}<{\centering}m{5pt}<{\centering}|m{75pt}<{\centering}|}
    \hline
   \multicolumn{7}{|c|}{\bf DecomposeNet: RGB encoder}\bigstrut\\
   \hline
   \textbf{Layer} & \textbf{K} & \textbf{S} & \textbf{Channels} & \textbf{I} & \textbf{O} & \textbf{Input Channels} \bigstrut\\
    \hline
    rgb\_conv0 & 6$\times$6   & 2     & 4/32    & 1 & 2 & $I_{\rm RGB}$, $I_{\rm D}$ \bigstrut[t]\\
    rgb\_conv1 & 4$\times$4  & 2     & 32/64   & 2 & 4 & rgb\_conv0 \\
    rgb\_conv2 & 4$\times$4  & 2  & 64/128  & 4 & 8 & rgb\_conv1 \\
    rgb\_conv3 & 4$\times$4   & 2     & 128/256 & 8 & 16 & rgb\_conv2\\
    rgb\_conv4 & 4$\times$4   & 2  & 256/512 & 16 & 32 & rgb\_conv3 \\

    \hline
    \multicolumn{7}{|c|}{\bf DecomposeNet: 3 copies of albedo, normal and roughness decoders}\bigstrut\\
   \hline
   
    upconv0 & 4$\times$4   & 2      & 512/256 & 32 & 16 & rgb\_conv4 \\
    upconv1 & 4$\times$4   & 2   & 256/128 & 16 & 8 & rgb\_conv3, upconv0 \\
    upconv2 & 4$\times$4   & 2      & 128/128 & 8 & 4 & rgb\_conv2, upconv1 \\
    upconv3 & 4$\times$4   & 2   & 128/64 & 4 & 2 & rgb\_conv1, upconv2 \\
    upconv4 & 4$\times$4   & 2      & 64/64  & 2 & 1 & rgb\_conv0, upconv3 \\
    \hline
    \multicolumn{7}{|c|}{\bf DecomposeNet: albedo, normal and roughness estimation layers}\bigstrut\\
    \hline
    Albedo & 5$\times$5   & 1   & 64/3   & 1 & 1 & upconv4 \\
    \hline
    Normal & 5$\times$5   & 1   & 64/3   & 1 & 1 & upconv4 \\
    \hline
    roughness & 5$\times$5   & 1   & 64/1   & 1 & 1 & upconv4 \\
    \hline
     \end{tabular}%
    \vspace{3pt}
\end{table}%

\begin{table}[htbp]
\footnotesize
  \centering
  \renewcommand{\arraystretch}{1}
   \begin{tabular}{|m{29pt}<{\centering}|m{8pt}<{\centering}m{3pt}<{\centering}m{27pt}<{\centering}|m{5pt}<{\centering}m{5pt}<{\centering}|m{75pt}<{\centering}|}
     \hline
    \multicolumn{7}{|c|}{\bf ShadowNet}\bigstrut\\
   \hline
   \textbf{Layer} & \textbf{K} & \textbf{S} & \textbf{Channels} & \textbf{I} & \textbf{O} & \textbf{Input Channels} \bigstrut\\
    \hline
    conv0 & 6$\times$6   & 2        & 3/32    & 1 & 2 & $T(\omega, T_{\rm D})$\\
    conv1 & 4$\times$4  & 2     & 32/64   & 2 & 4 & conv0 \\
    conv2 & 4$\times$4   & 2        & 64/128  & 4 & 8 & conv1 \\
    conv3 & 4$\times$4  & 2     & 128/256 & 8 & 16 & conv2 \\
    conv4   & 4$\times$4   & 2      & 256/256 & 16 & 32 & conv3  \\
    upconv0 & 4$\times$4   & 2     & 256/256 & 32 & 16 & conv4 \\
    upconv1 & 4$\times$4  & 2     & 256/256 & 16 & 8 & conv3, upconv0 \\
    upconv2 & 4$\times$4   & 2  & 256/128 & 8 & 4 & conv2, upconv1 \\
    upconv3 & 4$\times$4  & 2      & 128/64 & 4 & 2 & conv1, upconv2 \\
    upconv4 & 4$\times$4   & 2   & 64/32 & 2 & 1 & conv0, upconv3 \\
     \hline
    shadow & 6$\times$6   & 2      & 32/1 & 1 & 1 & upconv4 \\
    \hline
     \end{tabular}%
    \vspace{3pt}
\end{table}%

\begin{table}[htbp]
\footnotesize
  \centering
  \renewcommand{\arraystretch}{1}
   \begin{tabular}{|m{29pt}<{\centering}|m{8pt}<{\centering}m{3pt}<{\centering}m{27pt}<{\centering}|m{5pt}<{\centering}m{5pt}<{\centering}|m{75pt}<{\centering}|}
     \hline
    \multicolumn{7}{|c|}{\bf SynthesisNet}\bigstrut\\
   \hline
   \textbf{Layer} & \textbf{K} & \textbf{S} & \textbf{Channels} & \textbf{I} & \textbf{O} & \textbf{Input Channels} \bigstrut\\
    \hline
    conv0 & 6$\times$6   & 2        & 17/64    & 1 & 2 & $I_{\rm RGB}, T(\omega, T_{\rm D})$, \etc \\
    conv1 & 4$\times$4  & 2     & 64/128   & 2 & 4 & conv0 \\
    conv2 & 4$\times$4   & 2        & 128/128  & 4 & 8 & conv1 \\
    conv3 & 4$\times$4  & 2     & 128/256 & 8 & 16 & conv2 \\
    conv4   & 4$\times$4   & 2      & 256/256 & 16 & 32 & conv3  \\
    upconv0 & 4$\times$4   & 2     & 256/512 & 32 & 16 & conv4 \\
    upconv1 & 4$\times$4  & 2     & 512/256 & 16 & 8 & conv3, upconv0 \\
    upconv2 & 4$\times$4   & 2  & 256/128 & 8 & 4 & conv2, upconv1 \\
    upconv3 & 4$\times$4  & 2      & 128/64 & 4 & 2 & conv1, upconv2 \\
    upconv4 & 4$\times$4   & 2   & 64/32 & 2 & 1 & conv0, upconv3 \\
     \hline
    relight & 5$\times$5   & 2      & 32/3 & 1 & 1 & upconv4 \\
    \hline
    \end{tabular}%
    \vspace{3pt}
    \caption{Network architecture. {\bf K} means kernel size, {\bf S} means stride, and {\bf Channels} is the number of input and output channels. {\bf I} and {\bf O} are the input and output downsampling factor relative to the initial input. Separation by ``," in the {\bf Input Channels} means concatenation.}
  \label{tab:depth_net}%
\end{table}%
  
\section{BRDF model}
We use the microfacet BRDF model as in \cite{Karis2013RealSI}. In details, let $\mathbf{l}, \mathbf{v}$ be the unit vectors in the (negative of) incoming light direction and outgoing light direction, respectively. Let $\mathbf{h}$ be the unit vector in the same direction of $\frac{\mathbf{l} + \mathbf{v}}{2}$, and $\mathbf{n}$ be the surface normal vector. The BRDF model is written as
\[
f(\mathbf{l}, \mathbf{v}) = \frac{D(\mathbf{h})F(\mathbf{v}, \mathbf{h})G(\mathbf{l}, \mathbf{v})}{4 (\mathbf{n} \cdot \mathbf{v}) (\mathbf{n} \cdot \mathbf{v})}
\]
where
\[
D(\mathbf{h}) = \frac{\alpha^2}{\pi ((\mathbf{n}\cdot \mathbf{v})(\alpha^2 - 1) + 1)^2}
\]
here $\alpha = Roughness^2$, and
\[
G(\mathbf{l}, \mathbf{v}) = G_1(\mathbf{l})G_1(\mathbf{v})
\]
\[
G_1(\mathbf{v}) = \frac{\mathbf{n}\cdot \mathbf{v}}{(\mathbf{n}\cdot \mathbf{v})(1-k) + k }
\]
\[
k = \frac{(Roughness + 1)^2}{8}
\]
and
\[
F(\mathbf{v}, \mathbf{h}) = F_0 + (1 - F_0)^{((-5.55473(\mathbf{v} \cdot \mathbf{h})- 6.98316)(\mathbf{v} \cdot \mathbf{h})}
\]
where we use $F_0 = 0.05$.
\section{Additional experiments}

\subsection{Influence by environmental light}
An important issue in practice is that we often do not have a perfect dark room for capturing pure flash photographs, and it may not be always possible to subtract two photos of the same scene one with flash and one without. 
We therefore experiment with the effect of this additional environmental lighting component in the {\bf DecomposeNet} module. 
Specifically, we train a variant of our model, called {\tt Env}, with the following modifications. In the input, we replace the flashlight-only color image with a pair of images consist 1) a color image whose dominating light source is flashlight but also contains environmental light component, with background color coming from the environmental map, and 2) a binary mask that marks out the object to be relighted.
This is a much simplified design compared to {\tt SIPS}. 
Since estimating the environmental lighting is highly ill-posed, from Table \ref{tb:env} we see a drop in performance as compared to the {\tt Clean} baseline, however still compared favorably to {\tt SIPS}.

\begin{table}[h!]
\footnotesize
  \centering 
   \setlength{\tabcolsep}{2pt}
    \begin{tabular}{c|ccc}
    \hline
     MSE & Albedo & Normal & Roughness  \bigstrut\\
    \hline
    \rowcolor{Gray}  {\tt Env.} &  $\mathbf{0.99\times10^{-2}}$         &  $\mathbf{1.60\times10^{-2}}$    &    $\mathbf{4.47\times10^{-2}}$
    \bigstrut[t]\\
    \hline
     \cite{li2018learning} {\tt Init.} &   ${5.64\times10^{-2}}$      &  ${4.51\times10^{-2}}$     &  ${2.06\times10^{-1}}$   \bigstrut[t]\\
    \hline
     \rowcolor{Gray}   \cite{li2018learning} {\tt Refine.} & ${4.84\times10^{-2}}$  &  ${3.81\times10^{-2}}$   & ${1.93\times10^{-1}}$       \bigstrut[t]\\
    \hline
    \end{tabular}%
    \vspace{2pt}
    \caption{{\bf DecomposeNet} evaluation and comparison when input color image contains environmental lighting. Note that while our model is a single stage model, \cite{li2018learning} is a three-stage model with cascaded refinement modules. Here we show its results of initial estimation and final refinement.}
    \label{tb:env}
\end{table}%

\subsection{Result visualization video}
We have included a video to visualize the relighting under continuous change of the lighting directions as well as the environmental light. Specifically, we rotate the directional light source at $30$ degree about the optical axis, and rotate the environmental light source about optical axis. The results for both synthetic data and real data are shown.

\end{document}